\documentclass[10pt]{article}

\usepackage{amsmath}
\usepackage{amssymb}

\usepackage{graphicx}

\usepackage{cite}

\usepackage{color}

\usepackage{graphicx}
\usepackage{cite} 
\usepackage{url}  
\usepackage{ifthen}  
\usepackage{multicol}   
\usepackage{multirow}
\usepackage{subfigure}
\usepackage{epsfig}
\usepackage[labelfont=bf,labelsep=period,justification=raggedright]{caption}

\makeatletter
\renewcommand{\@biblabel}[1]{\quad#1.}
\makeatother

\date{}

\begin{document}

\begin{flushleft}
{\Large
\textbf{Juxtaposition of System Dynamics and Agent-based Simulation for a Case Study in Immunosenescence}
}
\\
Grazziela P Figueredo$^{\ast}$,
Peer-Olaf Siebers,
Uwe Aickelin,
Amanda Whitbrook,
Jonathan M Garibaldi
\\
\bf School of Computer Science, The University of Nottingham, Nottingham, United Kingdom
\\
$\ast$ E-mail: grazziela.figueredo@nottingham.ac.uk
\end{flushleft}

\section*{Author Summary}

Advances in healthcare and in the quality of life significantly increase human life expectancy. With the ageing of populations, new un-faced challenges are brought to science. The human body is naturally selected to be well-functioning until the age of reproduction to keep the species alive. However, as the lifespan extends, unseen problems due to the body deterioration emerge. There are several age-related diseases with no appropriate treatment; therefore, the complex ageing phenomena needs further understanding. Immunosenescence, the ageing of the immune system, is highly correlated to the negative effects of ageing, such as the increase of auto-inflammatory diseases and decrease in responsiveness to new diseases. Besides clinical and mathematical tools, we believe there is opportunity to further exploit simulation tools to understand immunosenescence. Compared to real-world experimentation, benefits include time and cost effectiveness due to the laborious, resource-intensiveness of the biological environment and the possibility of conducting experiments without ethic restrictions. Contrasted with mathematical models, simulation modelling is more suitable for representing complex systems and emergence. In addition, there is the belief that simulation models are easier to communicate in interdisciplinary contexts. Our work investigates the usefulness of simulations to understand immunosenescence by employing two different simulation methods, agent-based and system dynamics simulation, to a case study of immune cells depletion with age.

\section*{Abstract}

Advances in healthcare and in the quality of life significantly increase human life expectancy. With the ageing of populations, new un-faced challenges are brought to science. The human body is naturally selected to be well-functioning until the age of reproduction to keep the species alive. However, as the lifespan extends, unseen problems due to the body deterioration emerge. There are several age-related diseases with no appropriate treatment; therefore, the complex ageing phenomena needs further understanding. It is known that immunosenescence is highly correlated to the negative effects of ageing. In this work we advocate the use of simulation as a tool to assist the understanding of immune ageing phenomena. In particular, we are comparing system dynamics modelling and simulation (SDMS) and agent-based modelling and simulation (ABMS) for the case of age-related depletion of naive T cells in the organism.
We address the following research questions: Which simulation approach is more suitable for this problem? Can these approaches be employed interchangeably? Is there any benefit of using one approach compared to the other? Results show that both simulation outcomes closely fit the observed data and existing mathematical model; and the likely contribution of each of the naive T cell repertoire maintenance method can therefore be estimated.  The differences observed in the outcomes of both approaches are due to the probabilistic character of ABMS contrasted to SDMS. However, they do not interfere in the overall expected dynamics of the populations. In this case, therefore, they can be employed interchangeably, with SDMS being simpler to implement and taking less computational resources.

\section{Introduction}

Bulati {\it et al.}~\cite{Bulati:2008} characterises ageing as a complex process with negative repercussion on the development and functioning of the immune system. Progressive changes of the immune system components have a major impact on the capacity of an individual to produce effective immune responses. The decrease of immunocompetence in the elderly is the result of the continuous challenge and unavoidable exposure to a variety of potential antigens~\cite{Franceschi:2000}. Virus, bacteria, fungus, food, etc. cause persistent life-long antigenic stress, which is responsible for the filling of the immunological space by the accumulation of effector cells and immune memory.

Franceschi~\cite{Franceschi:2000} also identified some factors that characterise immune system ageing, many of which affect immune cells populations such as T cells. There is, for instance an accumulation of memory T cells and a marked reduction of the T cell repertoire. In addition, the thymic involution after twenties has a significant impact in the reduction of naive T cells numbers. Naive T cells are immunocompetent cells which role is to respond to new faced antigens. The depletion of these cells therefore eventually leaves the body more susceptible to new diseases~\cite{Murray:2003}.

This work is an extension of initial experiments with naive T cell populations presented in~\cite{Figueredo:2010}. Our objective is to study the decay of these cells numbers with age, under a simulation perspective.  We therefore compare SDMS and ABMS for an immune system ageing model involving interactions which influence the naive T cell populations over time. The model is based on a set of ordinary differential equations (ODEs) defined in~\cite{Murray:2003}. In their work, Murray {\it et al.} \cite{Murray:2003} propose a model with a set of equations to fit observed data and estimate the output of different naive T cells populations with age. We built equivalent models employing SDMS and ABMS, compare results and validate our outcomes with real-world data and the original ODE model~\cite{Murray:2003}. Our objective is to answer the following research questions: Which approach is more suitable for this problem? Can we use both approaches interchangeably? What are the benefits and problems encountered in each approach for this problem? In order to answer our research questions the naive depletion case study is investigated under five scenarios defined in~\cite{Murray:2003}. For each setting, the rates of changes within the cell populations are modified in order to determine how these changes are reflected under each simulation approach.

The remainder of this paper is organised as follows. Section~\ref{sec:Background} presents a review on the basic concepts of simulation, SDMS and ABMS. Section~\ref{sec:RelatedWork} introduces the literature concerned with the comparison of SDMS and ABMS. In Section~\ref{NaiveTCellsOutput} we describe the system used to define our simulations. Sections~\ref{sec:SD} and~\ref{sec:AB} introduce our SDMS and ABMS, respectively. Section~\ref{sec:Experiments} describes our experimental design, followed by the results obtained (Section~\ref{sec:Results}). Section~\ref{sec:Conclusions} presents our conclusions and opportunities for future work.

\subsection{Background}
\label{sec:Background}

A computational simulation of a system is as a ``{\it imitation (on a computer) of a system as it progresses through time}'' ~\cite{Robinson:2004}. Its purpose is to understand, change, manage and control reality~\cite{Pidd:2003}. Moreover, simulation is employed to obtain further understanding of the system and/or to identify improvements to a system~\cite{Robinson:2004}. A simulation predicts the performance of a system given a specific set of inputs. According to Robinson~\cite{Robinson:2004}, simulation is an experimental approach to modelling a ``{\it what-if}'' analysis tool. The model user determines the scenarios and the simulation predicts the outcomes. Simulation can, therefore, also be seen as a decision support tool. Compared to real-world experimentation, simulation is generally more cost-effective and less time consuming. Furthermore, under a controlled simulation environment, changes and different scenarios are analysed without impact to the real-world~\cite{Robinson:2004}, requiring no ethical agreements. The choice of the appropriate simulation method~\cite{Barton:2004,Lorenz:2009,Figueredo:2010} will determine the efficacy of the decision tool for a certain problem. In addition, different approaches impact on the content of the information produced by the simulation. Current major system simulation modelling methods consist of system dynamics modelling (SDM), discrete-event modelling, dynamic systems modelling and agent-based modelling (ABM)~\cite{Borshchev:2004}. We however investigate SDM and ABM, as they appear to be the most employed in immunology~\cite{Figueredo:2012:Thesis}.

System Dynamics (SD)~\cite{Forrester:1958} is an aspect of systems theory currently applied to any complex system characterized by interdependency, mutual interaction, information feedback and circular causality. The basis of the SD methodology is the recognition that the structure of a system is just as important in determining its behaviour as the individual components themselves. It is therefore necessary to adopt a ``systemic way of thinking''~\cite{Kirkwood:1998}, where the focus is on the top-down, internal system structure. This means that the problem should be depicted as a set of patterns, interrelated processes and generic structures~\cite{Pugh:1981}. SD works with feedback loops, stocks and flows that describe a system's nonlinearity. The changes that occur over time in the variables of the problem are ruled by ordinary differential equations. After understanding the structure of the problem to be simulated, it is necessary to translate it into a causal loop diagram, which is a graphical representation used in SD. Causal loop diagrams aid visualization of how interrelated variables affect one another. System Dynamics Simulation (SDS) is a continuous simulation for an SD model. It consists of a set of ordinary differential equations that are solved for a certain time interval~\cite{Macal:2010}. The simulation, therefore, has a deterministic output.

Agent-based modelling and simulation (ABMS) is a stochastic method, which employs a set of autonomous agents that interact with each other in a certain environment~\cite{Wooldridge:2002}. As it is derived from complex systems, its baseline is the notion that systems are built in a bottom-up perspective. In other words, an understanding of the dynamics of the system arises from individual interactions and their environment~\cite{Macal:2010b}. The agents' behaviours are described by rules that determine how they learn, interact with each other and adapt. The overall system behaviour is given by the agents' individual behaviours as well as by their interactions. ABMS is therefore well suited to modelling and simulating systems with heterogenous, autonomous and pro-active actors, such as human-centred systems, biological systems, businesses and organizations~\cite{Siebers:2007}. ABM can be done using state chart diagrams from the unified modelling language (UML). With state charts it is possible to define and visualize the agents' states, transitions between the states, events that trigger transitions, timing and agent actions~\cite{Borshchev:2004}.

The next section introduces the literature review concerned with the comparison between SDMS and ABMS for different problem domains. We initially review general work that has been carried out to assess the differences of both approaches. Subsequently, we focus on research concerned with the comparison of the strategies for immunological problems. We found, however, that there is a scarcity of literature comparing the two approaches for immune simulation.

\subsection{Related Work}
\label{sec:RelatedWork}

Scholl \cite{Scholl:2001a} conducts one of the first studies characterising the domains of applicability of SDMS and ABMS. The author also investigates the strengths and weaknesses of each approach, outlining opportunities for SDMS and ABMS-based multi-method modelling. Pourdehnad {\it et al.} \cite{Pourdehnad:2002} extends this pioneer work by comparing the two approaches conceptually. The authors discuss the potential synergy between both paradigms to solve problems of teaching decision-making processes. Similarly, Stemate { \it et al.} \cite{Stemate:2007} compare these modelling approaches and identify a list of likely opportunities for cross-fertilization. The authors see this list as a starting point for other researchers to take such synergistic views further.

Schieritz \cite{Schieritz:2002} and  Scheritz {\it et al.}~\cite{Schieritz:2003} contribute largely to the comparison of SDMS and ABMS for operational research (OR). They identify the unique features of each approach and present a table with their main differences. Further in \cite{Schieritz:2003b} the authors describe an approach to combine ODEs and ABMS for solving supply chain management problems. Their results show that the combined SDMS/ABMS does not produce the same outcomes as those from the SDMS alone. To understand why these differences occur, the authors indicate that more research needs to be conducted.

Ramandad {\it et al.} \cite{Sterman:2008} compare the dynamics of a ABMS and SDMS for contagious disease spread. The authors convert the ABM into an SDM and examine the impact of individual heterogeneity and different network topologies. They conclude that the deterministic SDMS yields a single trajectory for each parameter set, while the stochastic ABMS yield a distribution of outcomes. Moreover, the outcomes differ in several metrics relevant to public health.

Schieritz \cite{Schieritz:2004} analyses two arguments given in literature to explain the superiority of ABMS compared with SDMS for social simulation: (1) the inability of SD to represent emergence and (2) SD's lack of individual diversity.  The author points out that an agent-based approach models individuals and interactions on a lower level, implicitly taking up an individualist position of emergence; conversely, SD models social phenomena at an aggregate level. As a second part of the study, the author compares  SDM and ABM for modelling species competing for resources to analyse the effects of evolution on population dynamics. The conclusion is that when individual diversity is considered, it limits the applicability of the ODE model. However, it is shown that ``{\it a highly aggregate more ODE-like model of an evolutionary process displays similar results to the ABMS}''.

Similarly, Lorenz ~\cite{Lorenz:2009} proposes three aspects to be compared for the choice between SDMS and ABMS: structure, behaviour and emergence. Structure is related to how the model is built. The structure of a SDMS model is static, whereas in ABMS it is dynamic. In SDMS, all the elements of the simulation are developed in advance. In ABMS, on the other hand, agents are created or destroyed and interactions are defined through the course of the simulation run. The second aspect (behaviour) focuses on the central generators of behaviours in the model. For SDMS the behaviour generators are feedback and accumulations, while for ABM they are the interactions of the systems elements. Both methodologies incorporate feedback. ABMS, however, has feedback in more than one level of modelling. The third aspect lies in their capacity to capture emergence, which differs between the two methodologies. In disagreement with Schieritz~\cite{Schieritz:2004} mentioned earlier, the author states that ABMS is capable of capturing emergence, while the one-level structure of SDMS is insufficient in that respect.

In previous work~\cite{Figueredo:2012:Thesis} we discuss the merits of SDMS and ABMS for problems involving the interactions with the immune system and early-stage cancer. Our interest was to identify those cases in immunology where ABMS and SDMS  could be applied interchangeably without compromising the results usefulness. In addition, we wanted to investigate those circumstances where one approach was not able to reflect the compared approach outcomes. We found several outcome differences: (i) not everything  produced by the SDMS can be observed in ABMS outcomes (e.g. no half agents), which can impact in the outcomes similarity when the populations size is small (less than 50 agents); (ii) for our case studies, ABMS contributed to additional insights; due to its stochastic nature and emergent behaviour it produced different results and extra patterns of behaviour. In this work we continue our investigations by applying the comparison to an immunosenescence problem, as shown next.

\section{Methods}

\subsection{Case Study: Naive T Cells Output}
\label{NaiveTCellsOutput}

Before the age of 20, the set of naive T cells is sustained primarily by the thymus~\cite{Murray:2003}. Thymic contributions in an individual are quantified by the presence of a biological marker known as `T cell receptors excision circle' (TREC), which is a circular DNA originated during the formation of the T-cell receptor. The percentage of T cells with TRECs decays with shrinkage of thymic output, activation and reproduction of naive T cells~\cite{Murray:2003}. This means that naive T cells from the thymus have a greater percentage of TREC than those originating from other sources. In middle age, however, the thymus involutes and there is a change in the source of naive T cells. There is a considerable reduction in thymic naive T cell output, which means that new naive T cells are mainly produced by existing cells reproduction (peripheral expansion). It is believed that the naive T cell repertoire after twenties is also maintained by the population of existing memory cells, which have their phenotype reverted back to the naive cells type~\cite{Murray:2003}.

These two new methods of naive T cell repertoire maintenance, however, are insufficient to keep an effective defense system in the organism~\cite{Murray:2003}, as they do not produce new phenotypic changes in the T cells. Rather, evidence shows that they continue to fill the naive T cell space with copies of existing cells~\cite{Franceschi:2005}, which are incapable to eliminate new antigens. The loss (death) of clones of some antigen-specific T cells therefore becomes irreversible. These age-related phenomena lead to a decay of immune performance.

Our study on the dynamics of naive T cells over time is based on equations obtained in~\cite{Murray:2003} and real-world data from~\cite{Murray:2003,Cossarizza1996173,Lorenzi:2008}. The model objective is to investigate the likely contribution of each of the naive T cell's sources by comparing estimates of the presence of TREC in the cells. In the model, four populations are considered, naive T cells from thymus, naive T cells from peripheral proliferation, active cells and memory cells. The mathematical model of the dynamics of the cell populations and its sustaining sources is presented next.

\subsubsection{The Mathematical Model}
\label{cap:CaseStudy1:MathematicalModelNaiveOutput}

The model proposed in~\cite{Murray:2003} is described by equations~\ref{eq:modelNaiveTCell:equation1} to~\ref{eq:modelNaiveTCell:equation6}, in which $N$ is the number of naive cells from thymus, $N_{p}$ is the number of naive cells that have undergone proliferation, $A$ are the active cells, $M$ is the number of memory cells and $t$ is time (in years). At the beginning of life most naive T cells belong to the population $N$. With time naive T cells from thymus proliferate, which contributes for the increase of the $N_{p}$ population. When the body faces a new threat, naive T cells are recruited and become active ($A$). A fraction of active cells turns into memory cells ($M$).

The first differential equation describing the naive T cell population from thymus is:

\begin{equation}
\frac{dN}{dt} = s_0e^{-\lambda_tt}s(N_p)-[\lambda_n+\mu_ng(N_p)]N
\label{eq:modelNaiveTCell:equation1}
\end{equation}

where $s_0$ is the thymic output value, $\lambda_t$ is the thymic decay rate, $t$ represents time in years, $s_0e^{-\lambda_tt}s(N_p)$ is the number of cells that arise from the thymus and $s(N_p)$ is the rate of export of the thymus defined by:

\begin{equation}
s(N_p) = \frac{1}{1+\frac{\bar{s}^{N_p}}{\bar{N_p}}}
\label{eq:modelNaiveTCell:equation2}
\end{equation}

$\bar{N_p}$ and $\bar{s}$ are equilibrium and scaling values respectively. These values were established in~\cite{Murray:2003}. $\lambda_nN$ represents the naive cells that become part of the naive proliferating population, $\lambda_n$ is the naive proliferation rate, $\mu_n$ is the thymic naive cells death rate, $\mu_ng(N_p)N$ represents the naive cell death rate and $g(N_p)$ is the death rate between naive TREC-positive and naive TREC-negative cells, defined as:

\begin{equation}
g(N_p) = 1+ \frac{\frac{bN_p}{\bar{N_p}}}{1+\frac{N_p}{\bar{N_p}}}
\label{eq:modelNaiveTCell:equation3}
\end{equation}

The second differential equation describing the naive T cells from proliferation is:

\begin{equation}
\frac{dN_p}{dt} = \lambda_nN + [ch(N,N_p) - \mu_n]N_p + \lambda_{mn}M
\label{eq:modelNaiveTCell:equation4}
\end{equation}

where $c$ is the proliferation rate, $ch(N,N_p)N_p$ represents the naive proliferation and $h(N,N_p)$ is the dilution of thymic-naive through proliferation defined by:

\begin{equation}
h(N,N_p) = \frac{1}{1+\frac{N+N_p}{\bar{N_p}}}
\label{eq:modelNaiveTCell:equation5}
\end{equation}

$\mu_nN_p$ is the death rate of proliferation-originated naive cells and $\lambda_{mn}$ is the reversion rate from memory into $N_p$.

The final differential equation for the memory cell population dynamics is:

\begin{equation}
\frac{dM}{dt} = \lambda_aA - \mu_mM - \lambda_{mn}M
\label{eq:modelNaiveTCell:equation6}
\end{equation}

where $\lambda_a$ is the reversion rate into memory and $\mu_m$ is the death rate of memory cells.

\begin{figure}[ht]
 \begin{center}
  \resizebox{11cm}{!}{\includegraphics{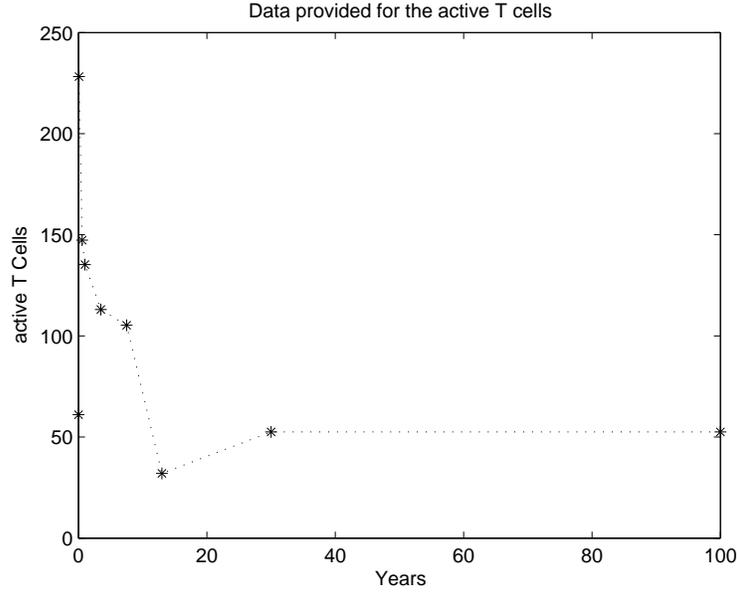}}
 \end{center}
 \caption{The data set used as a look-up table for the active cells}
  \label{fig:ModelNaiveTCellDataActive}
\end{figure}

The parameter values for the model are shown in Table~\ref{Tab:RatesModelNaiveTCell}. For the mathematical model and subsequent simulations, $s0 =56615$. The values for active cells over time are determined by referring to data collected by \cite{Comans-Bitter:1997} (Figure~\ref{fig:ModelNaiveTCellDataActive}). This table contains the number of activated CD4+ cells (a type of naive T cells) per $mm^3$ for early years and is used as a stock for the active cells.
From the active cell stock the values of the memory cell stock are updated according to the parameter $\lambda_{a}$.

Equations~\ref{eq:modelNaiveTCell:equation1}  to \ref{eq:modelNaiveTCell:equation6} are incorporated in the SDM and ABM in order to investigate if it is possible to reproduce and
validate the results obtained in~\cite{Murray:2003}. Moreover, variations of the ratio variables are explored to understand the importance of each individual integrand in the system. For example, it is important to establish how much the proliferation rate impacts on the depletion of naive T cells over age, and to identify the point in time at which the system can be defined as losing functionality.

\subsection{The System Dynamics Modelling and Simulation}
\label{sec:SD}

\begin{figure}[ht]
  \centering
    \resizebox{14cm}{!}{\includegraphics{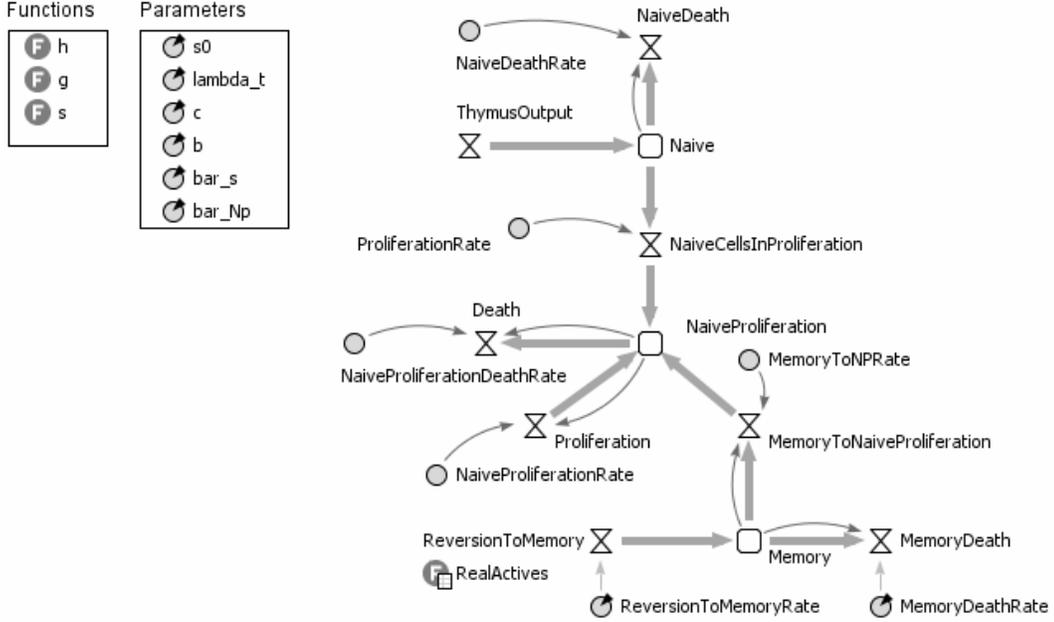}}
  \caption{The system dynamics model, functions and parameters}
  \label{fig:ModelNaiveTCellsSD}
\end{figure}

The SD's model stock and flow diagram is constituted by the stocks, flows, information, auxiliaries and parameters, as shown in Figure~\ref{fig:ModelNaiveTCellsSD}. In the model, the naive T cells, naive T cells from proliferation and memory cells are the stock variables, as the aim is to keep information of how they accumulate over time.

In the figure, the stock variable that represents the number of naive T cells ($Naive$) is subject to inflowing thymic output ($ThymusOutput$), and proliferation ($NaiveCellsInProliferation$) and death ($NaiveDeath$) outflows. The stock of naive cells from proliferation's ($NaiveProliferation$) inflows are proliferation and reversion from memory ($MemoryToNaiveProliferation$), and the outflow is death, according to Equation~\ref{eq:modelNaiveTCell:equation2}.  The memory stock's inflow is reversion from active ($ReversionToMemory$). The outflows are reversion to a naive phenotype (in the figure, {\it MemoryToNaiveProliferation} and death, as defined by Equation~\ref{eq:modelNaiveTCell:equation6}). The number of active cells, which is a stock, is given by real-world data of active cells in the human organism (in the figure, it is the table $RealActives$).

Information (curved arrows in the stock and flow diagram) between stocks and flows indicates that there is an information about a stock that influences a flow. By looking at Equation~\ref{eq:modelNaiveTCell:equation1}, it is possible to determine that there is information between the stock $Naive$ ($N$ in Equation~\ref{eq:modelNaiveTCell:equation1}) and the flow $NaiveDeath$. For the $NaiveProliferation$ stock there is information from it to $Proliferation$ and $Death$ flows. In the $Memory$ stock there is information from it to the flows {\it MemoryToNaiveProliferation} and $MemoryDeath$. The model parameters are the same as those in the mathematical model.

For our implementation, functions are designed for $s$ and $g$, which use the stock $NaiveProliferation$ in their calculations. Hence, the information about $NaiveProliferation$ is implicit in these functions.

The mathematical parameters and their correspondents in the SD model are shown in Table~\ref{Tab:ParametersModelNaiveTCell}.

Table~\ref{Tab:NaiveFlowCalculations} demonstrates the flows for each stock, their correspondent in the mathematical model and the flow formula. In the table, the functions $s0()$, $s()$, $g()$ and $h()$ are implemented according to corresponding mathematical functions. The function $time()$ returns the current simulation time, which, for this case, is given in years. Furthermore, the $ThymusOuput$ is an example of flow which does not have any information or parameter. Hence, it is defined according to the mathematical expression stated.

\subsection{The Agent-based Modelling and Simulation}
\label{sec:AB}

\begin{figure}[!ht]
 \begin{center}
  \resizebox{14cm}{!}{\includegraphics{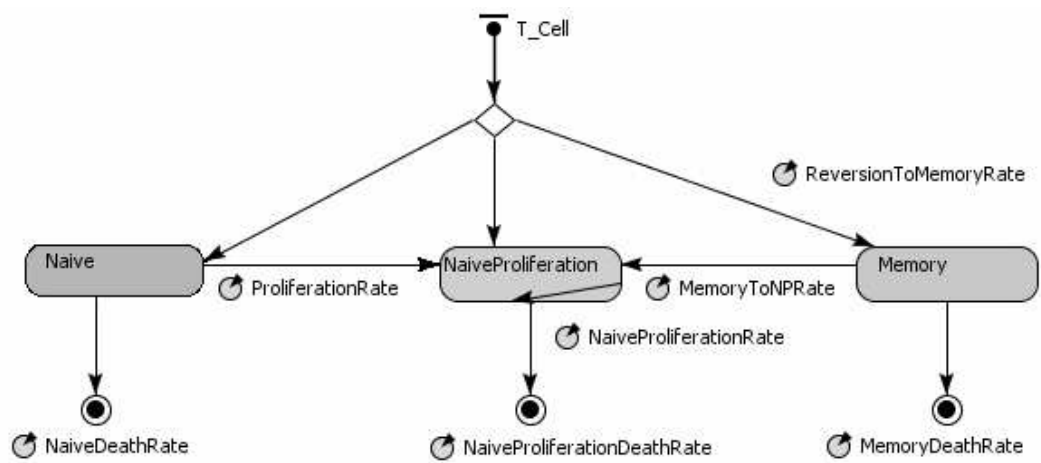}}
 \end{center}
 \caption{The naive T cell agent}
 \label{fig:ModelNaiveTCellsABS}
\end{figure}

In our model T cells are the agents, which can assume three states: $Naive$, $NaiveFromProliferation$ and $Memory$, as shown in the state chart depicted in Figure~\ref{fig:ModelNaiveTCellsABS}. The lozenge in the top of the state chart represents a branch for the decision of the T cell current state. In the figure, there are also final states when cells die and are eliminated from the system (represented by a black circle outlined with a smaller filled in circle inside it). Each agent behaviour depends on its current state and occurs according to a certain parameter rate. The agent's parameters and behaviours corresponding to each state are shown in Table~\ref{Tab:NaiveABSBehaviours}.

The agents' transitions (arrows) determine the changes in states. The state changes and death rates are given by the ratios defined in the mathematical model. Initially, all the agents are in the {\it naive} state. As the simulation proceeds, they can assume other stages according to the transition pathways defined in the state chart.

\begin{figure}[!ht]
 \begin{center}
  \resizebox{12cm}{!}{\includegraphics{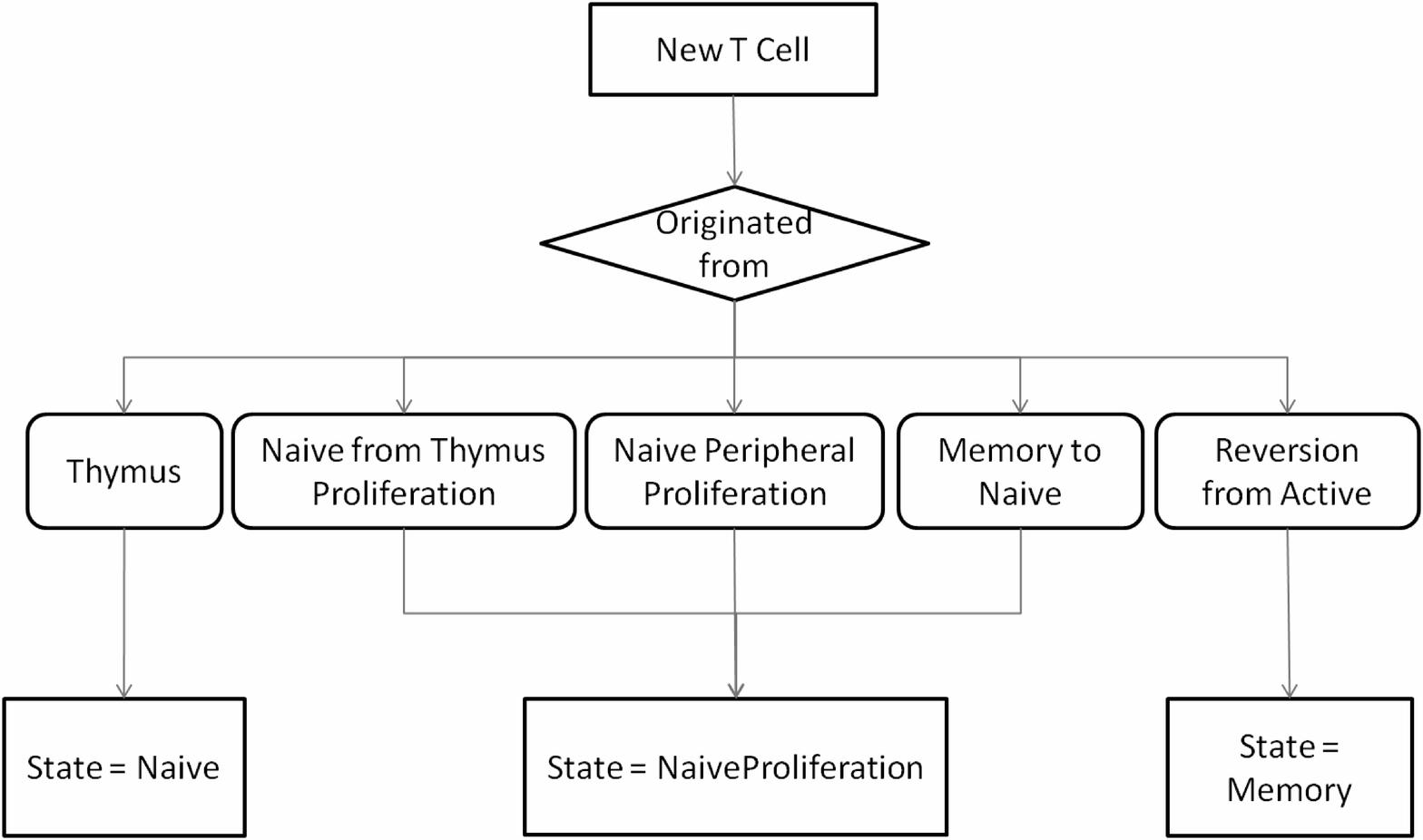}}
 \end{center}
 \caption{New agent (T cell) state decision flow chart}
 \label{fig:FlowChartStatesDefinition}
\end{figure}

When agents reproduce, the newborn agents, which are also T cells, should assume the same state as their original agent. Apart from proliferation, new agents are also produced from thymic output and reversion from active to memory cells. The algorithm that determines the agent state is given according to the flow chart in Figure~\ref{fig:FlowChartStatesDefinition}.

Our agents respond to changes in time and do not interact with each other directly. For the simulation development, apart from the agents, there is also a function that determines the thymic output and the number of active cells (from the look-up table) that become memory cells. Both are implemented using events that determine when each of these T cells should enter the system. The thymic output calculation, the functions $s$, $g$, $h$  and the active cells look-up table are the same as those from the SD model.

\subsection{Experiments}
\label{sec:Experiments}

Five simulation scenarios were studied, defined by~\cite{Murray:2003} with different values for the parameters. A summary of parameters changed for each scenario is illustrated in Table~\ref{Tab:ModelNaiveTCellsParameters}.

The first scenario investigates whether there is the need for naive peripheral proliferation throughout life to sustain the naive population. The naive peripheral proliferation rate for this experiment is therefore set to zero. It also considers reversion from memory to a naive phenotype.

The second scenario assumes peripheral proliferation with a higher rate of naive cells becoming naive proliferating cells $(\lambda_n = 2.1)$. There is no reversion from memory to a naive phenotype and no homeostatic reduction in thymic export. The functions $s$, $g$ and $h$ from the mathematical model are responsible for controlling the thymic export, naive death rate and naive peripheral proliferation respectively. In order to alter the thymic export, the parameters $\overline{s}$ and $\overline{N}_p$ are changed. The parameter $b$ is set to zero so that the function $g$ remains constant during the entire simulation, as does the death rate of naive cells.

The third scenario alters the function $g$ over time by setting the parameter $b$ greater than zero ($b=4.2$). This means that the death rate of naive T cells from thymus increases along the years as the number of naive from peripheral proliferations rises. There is no change to the thymic export, no reversion from memory to a naive phenotype and the conversion rate of naive from thymus to naive proliferation is low (equal to 0.003).

Scenario 4 intends to produce the opposite results from those of scenario 3. In this case there is no change in the death rate of naive T cells from thymus. Rather, there is change on the thymic export with time.

Finally, the fifth scenario presents no restrictions, which means that there are changes in thymic export and death of naive cells over time. Moreover, there is peripheral proliferation and no memory turns back to a naive phenotype.

Two datasets were used for validation of the simulations. They are displayed in
Tables~\ref{Tab:datasetMurray} (obtained from\cite{Murray:2003} and \cite{Cossarizza1996173}) and~\ref{Tab:datasetLorenzi} (obtained from~\cite{Lorenzi:2008}). The datasets contain information about the TREC marker in individuals grouped in age ranges. In the tables, the first column shows the age range of the individuals; the second column has the mean $\frac{\log_{10}TREC}{{10^6}} PBMC$ (peripheral blood mononuclear cell) and the third column contains the number of individuals in each age range.

\begin{figure}[!ht]
 \begin{center}
  \resizebox{14cm}{!}{\includegraphics{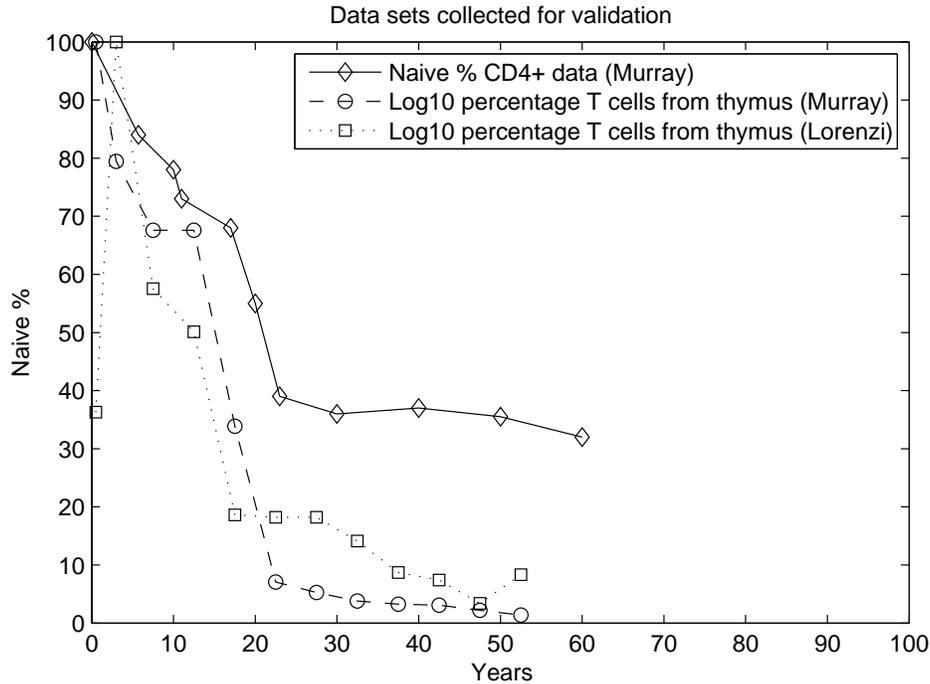}}
 \end{center}
 \caption{Data sets (collected in \cite{Murray:2003}, \cite{Cossarizza1996173} and \cite{Lorenzi:2008}) used for validation of the naive T cell output simulation models}
 \label{fig:ModelNaiveTCellsTwoDataSets}
\end{figure}

The diagram containing the TREC data (naive from thymus) and total naive cell data provided by~\cite{Murray:2003} \cite{Cossarizza1996173} and \cite{Lorenzi:2008} is shown in Figure~\ref{fig:ModelNaiveTCellsTwoDataSets}. In the figure, data provided in Table~\ref{Tab:datasetMurray} is represented by the symbol $\bigcirc$ ; the $ \square$ symbol indicates the data from Table~\ref{Tab:datasetLorenzi}. In addition, the total percentage of naive T cells in the body, obtained in~\cite{Murray:2003}, is also displayed (symbol $\lozenge$).

Each simulation was run for a period of one hundred years taking into account the impact of thymic shrinkage per $mm^3$ of peripheral blood and $3673$ initial naive cells from thymus. For the ABM, the simulation was run fifty times and the mean result of these runs was collected.

\section{Results}
\label{sec:Results}

\begin{figure}
\centering
\subfigure[ODE]{
   \includegraphics[scale =0.6] {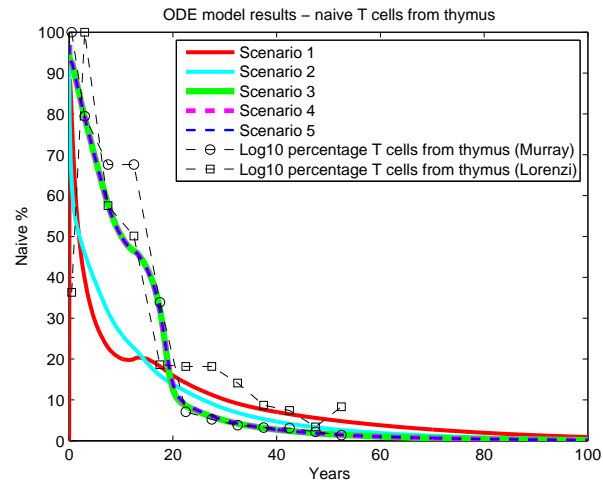}
   \label{fig:ModelNaiveTCellsNaiveThymusSDS}
 }

\subfigure[SDS]{
   \includegraphics[scale =0.6] {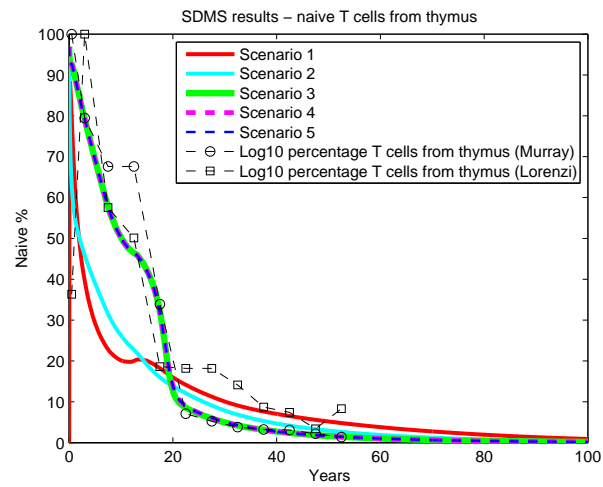}
   \label{fig:ModelNaiveTCellsNaiveThymusSDS}
 }

 \subfigure[ABMS]{
   \centering
   \includegraphics[scale =0.6] {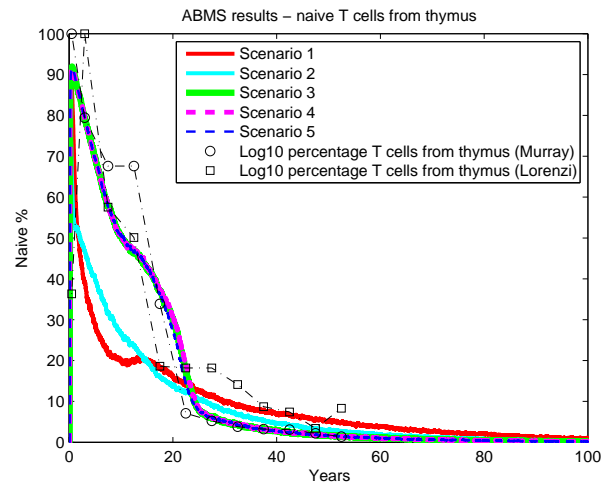}
   \label{fig:ModelNaiveTCellsNaiveThymusABMS}
 }
\caption{Results for naive T cells from thymus}
\label{fig:ModelNaiveTCellsNaiveOutput}
\end{figure}

\begin{figure}
\centering
\subfigure[ODE]{
   \includegraphics[scale =0.6] {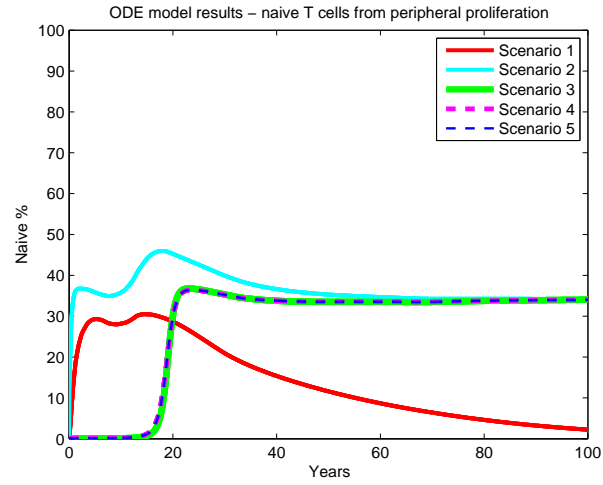}
   \label{fig:ModelNaiveTCellsPeripheralProliferationSD}
 }
\subfigure[SDS]{
   \includegraphics[scale =0.6] {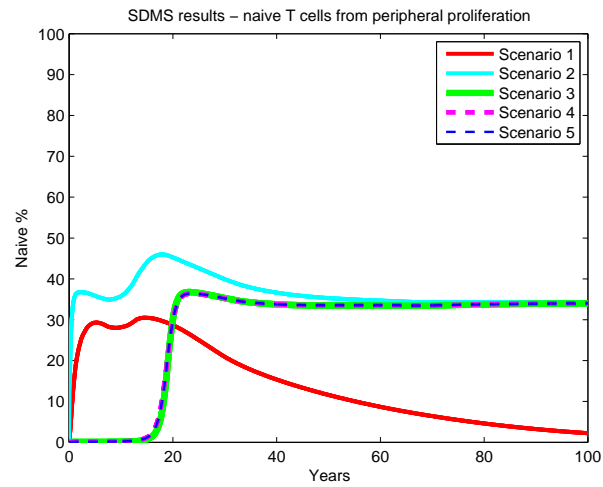}
   \label{fig:ModelNaiveTCellsPeripheralProliferationSD}
 }
 \subfigure[ABMS]{
   \includegraphics[scale =0.6] {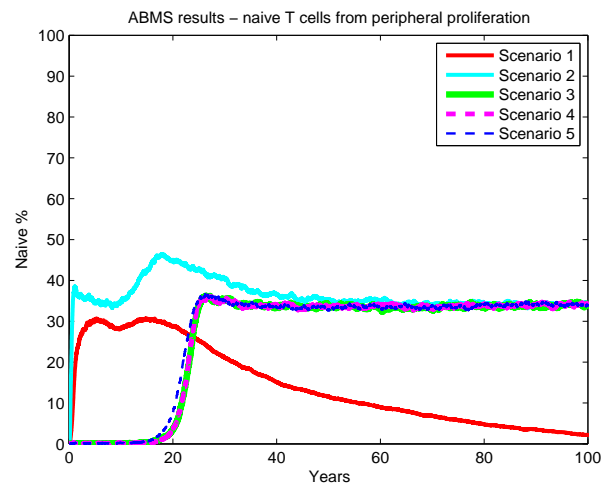}
   \label{fig:ModelNaiveTCellsPeripheralProliferationABMS}
 }
\caption{Results for naive T cells from peripheral proliferation}
\label{fig:ModelNaiveTCellsPeripheralProliferation}
\end{figure}

\begin{figure}
\centering
\subfigure[ODE]{
   \includegraphics[scale =0.6] {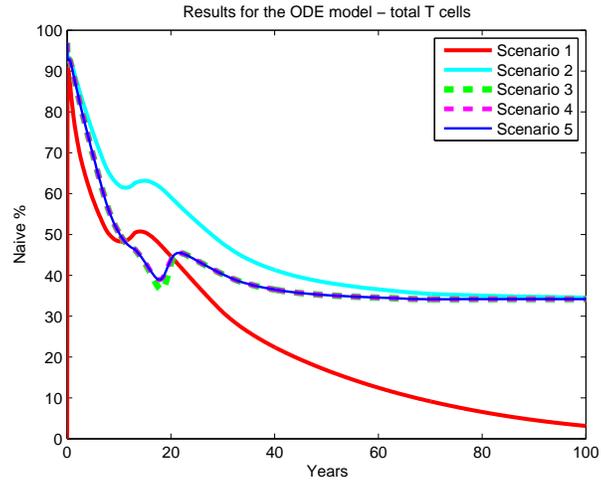}
   \label{fig:ModelNaiveTCellsTotalSD}
 }
\subfigure[SDMS]{
   \includegraphics[scale =0.6] {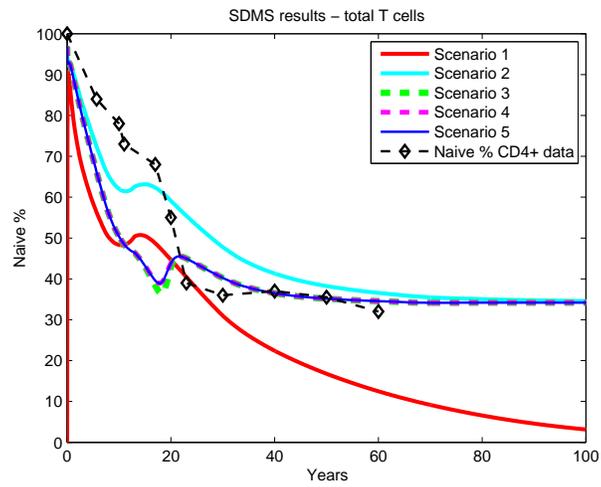}
   \label{fig:ModelNaiveTCellsTotalSD}
 }
 \subfigure[ABMS]{
   \includegraphics[scale =0.6] {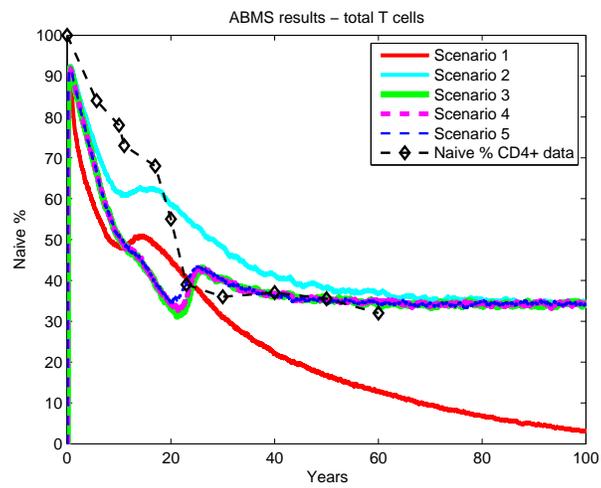}
   \label{fig:ModelNaiveTCellsTotalABMS}
 }
\caption{Results for total T cells}
\label{fig:ModelNaiveTCellsTotal}
\end{figure}

The simulation results contrasting SDMS with ABMS are illustrated in Figures~\ref{fig:ModelNaiveTCellsNaiveOutput}, \ref{fig:ModelNaiveTCellsPeripheralProliferation} and \ref{fig:ModelNaiveTCellsTotal}. Figures~\ref{fig:ModelNaiveTCellsNaiveOutput}(a), \ref{fig:ModelNaiveTCellsPeripheralProliferation}(a) and \ref{fig:ModelNaiveTCellsTotal}(a) show the ODE results used as a baseline for our results validation. Overall, when comparing SDMS and ABMS outputs, the results were similar. As expected, the ABMS produced some variation on the simulation curves while SDMS's curve was steady. In addition, SDMS took far less computational resources.

In the first scenario, the results for both simulation techniques show a very similar trend curve, although the ABMS results exhibit a more noisy behaviour in time. Results did not fit the original data (Tables~\ref{Tab:datasetMurray} and~\ref{Tab:datasetLorenzi}). We believe, however that the ABMS results can be improved to match the data more closely. We employed the same laws and rates from the original ODE model, which is one of the shortcomings of our work. We intend in the future to calibrate the ABMS solely with the data provided and observe the changes in the outcome.

Regarding the biological dynamics, naive cells from thymus curve demonstrated a substantial decay in thymic export on the beginning of life because of the high death rate. After the twenties, an exponential decay of thymic export was observed and the dynamics followed the thymic decay rate rule defined in the mathematical model. The naive proliferation curve increased with the decrease of naive from thymus, but as there was no proliferation of peripheral cells, they died with no replacement. Thus they followed the same pattern as that of their only source, i.e. thymic naive cells. The results indicate that peripheral proliferation is important for maintenance of naive T cells.

Results from scenario $2$ matched the original data more closely for both approaches. This case considered peripheral proliferation, as well as a high rate of naive cells from the thymus turning into peripheral naive cells. The naive from thymus curve shows a substantial decay in the beginning of life because of the death and proliferation rates. On the other hand, the naive from proliferation curve increased with the decrease of the naive from thymus curve. The main difference between these results and the results from the previous scenario is that the number of naive cells from proliferation reached a stable value after the age of twenty with no further decay. The results indicate the importance of peripheral expansion, but also the need for a smaller rate of naive to peripheral naive conversion. Moreover, reversion from memory to a naive phenotype does not seem to influence the overall quantities of cells.

Scenario $3$ took into account the results produced in the previous scenarios and adjusted the parameters in a way that a more accurate output was obtained. The naive from thymus curve presented a decay at the beginning of life followed by an interval of stability. By the age of twenty the thymic export decreased in an exponential trend. With the decay of naive from thymus, the naive repertoire changed from the thymic source to peripheral proliferation source. By performing these simulations it is therefore possible to have an idea of how the decay of naive cells occurs over time. The results now closely matched the original data. Scenarios $4$ and $5$ produced similar results to scenario 3. This indicates that alterations in thymic export and in naive death do not interfere significantly with the overall dynamics of the naive T cells.

In the five scenarios studied, the simulations produced similar results for both SDMS and ABMS. This can also be observed in the results of Wilcoxon rank sum tests applied to both ABMS and SDMS
results (Table~\ref{Tab:ModeNaiveTCellsWilcoxon}). The table reports p-values associated with Wilcoxon rank sum tests for the five scenarios. Our hypothesis is that the outcomes produced are not significantly different. The p-values for each test all exceed the 0.05 (5\%) significance level, indicating that the distributions of the outcomes of the various simulation approaches are not statistically different and therefore, the tests failed to reject the similarity hypothesis.

\section{Discussion}
\label{sec:Conclusions}

Besides mathematics, simulation becomes more and more popular in immunology research; however, comparisons of different simulation methods in this field are very rare. In this research we contrasted two different simulation methods, SDMS and ABMS, for an immunosenescence case study and tested if they provided us with a different insight. Our simulation models were based on mathematical equations describing thymic decay and naïve cells dynamics converted into SDMS and ABMS. Five simulation scenarios were studied investigating different sustaining sources for the naive cell population. Our research questions were: Which simulation approach is more suitable for this problem? Can these approaches be employed interchangeably? Is there any benefit of using one approach compared to the other?

Results for both methods matched those from the original ODE model. In the ABMS simulation, cells were subject to individual rates that occurred during the time slot in which they were created. However, this did not seem to have interfered in the final outcome. Further, as our experiments involved large populations, the variability inherent in ABMS did not considerably affect the overall dynamics of the simulated population. In the SDMS simulation outputs were exactly the same as those from the ODEs, as expected. The SDMS was simpler to implement and required significantly less computational resources such as memory, processing time and complexity. Therefore, although in this work both approaches can be employed interchangeably, SDMS seems more suitable.

The SDM and ABM were built based on the initial mathematical rules used to fit of the data collected, which is a limitation of our work. This induces the output of the simulations to behave similarly to that from the ODE model. The new results therefore were no more informative than the original ODE model, as they fit the data in the same manner. Furthermore, incorporating the functions for thymus output, g, s and h to the ABM made it hybrid; and its stochastic character did not produce any significant variant (extreme pattern) of the expected outcomes. Emergence was also not observed.

Another relevant aspect is that the agents considered are static (no movement) and non-interacting, which was a limitation imposed by the way the biological system was described. The features of the problem studied therefore did not allow us to exploit the ABMS capabilities appropriately.

The advantage of applying SDM and ABM to this case study is therefore mostly due to the fact that the modelling processes are more intuitive than the original ODE model. Additionally, the diagrams of both SDM and ABM are easier to communicate in multidisciplinary contexts. In particular, due to the characteristics of the problem, SDM also appears  to be more suitable for the modelling process. We believe therefore that the value of employing simulation in this case study context is in model communication to non-experts. Our future goal is to find ways to formalise the translation from ODE to SDM and ABM and to find out which of these is actually preferred by practitioners.

As another future aim, to overcome our modelling and simulation shortcomings it is our intention to rebuilt and calibrate our models based solely on the data provided. We intend to assess and report on the impacts of this procedure. We hypothesize that changes in the SDMS and ABMS can show further differences in the results. In addition, we believe we can achieve a better representation of the biological phenomena and therefore better fitting of the data. We intend to subsequently validate our modelling efforts with imunologists. This should help enforcing the idea regarding the usefulness of SDM and ABM simulations to understand immunosenescence, contrasted to the traditional mathematical modelling.

\section*{Acknowledgements}

We would like to thank Dr. Alice R. Lorenzi and Professor John D. Isaacs for providing us one of the data sets for our experiments.
We would also like to thank Dr. John Murray and D. D. Ho for access to their data set and help with this work.

\bibliography{tese}

\begin{thebibliography}{10}
\providecommand{\url}[1]{\texttt{#1}}
\providecommand{\urlprefix}{URL }
\expandafter\ifx\csname urlstyle\endcsname\relax
  \providecommand{\doi}[1]{doi:\discretionary{}{}{}#1}\else
  \providecommand{\doi}{doi:\discretionary{}{}{}\begingroup
  \urlstyle{rm}\Url}\fi
\providecommand{\bibAnnoteFile}[1]{%
  \IfFileExists{#1}{\begin{quotation}\noindent\textsc{Key:} #1\\
  \textsc{Annotation:}\ \input{#1}\end{quotation}}{}}
\providecommand{\bibAnnote}[2]{%
  \begin{quotation}\noindent\textsc{Key:} #1\\
  \textsc{Annotation:}\ #2\end{quotation}}
\providecommand{\eprint}[2][]{\url{#2}}

\bibitem{Bulati:2008}
Bulatti M, Pellicanò M, Vasto S, Colonna-Romano G (2008) Understanding ageing:
  Biomedical and bioengineering approaches, the immunologic view.
\newblock Immunity \& Ageing 5.
\bibAnnoteFile{Bulati:2008}

\bibitem{Franceschi:2000}
Franceschi C, Bonafè M, Valensin S (2000) Human immonosenescence: the
  prevailing of innate immunity, the failing of clonotypic immunity, and the
  filling of immunological space.
\newblock Vaccine 18: 1717-1720.
\bibAnnoteFile{Franceschi:2000}

\bibitem{Murray:2003}
Murray JM, Kaufmann GR, Hodgkin PD, Lewin SR, Kelleher AD, et~al. (2003) Naive
  {T} cells are maintained by thymic output in early ages but by proliferation
  without phenotypic change after twenty.
\newblock Immunology and Cell Biology 81: 487-495.
\bibAnnoteFile{Murray:2003}

\bibitem{Figueredo:2010}
Figueredo GP, Aickelin U (2010) Investigating immune system aging: System
  dynamics and agent-based modelling.
\newblock In: Proceedings of the Summer Computer Simulation Conference 2010.
\bibAnnoteFile{Figueredo:2010}

\bibitem{Robinson:2004}
Robinson S (2004) Simulation: The Practice of Model Development and Use.
\newblock John Wiley and sons, Ltd.
\bibAnnoteFile{Robinson:2004}

\bibitem{Pidd:2003}
Pidd M (2003) Tools for thinking: Modelling in management science.
\newblock Wiley. Chichester, UK.
\bibAnnoteFile{Pidd:2003}

\bibitem{Barton:2004}
Barton P, Brian S, Robinson S (2004) Modelling in the economic evaluation of
  health care: selecting the appropriate approach.
\newblock Journal of Health Services Research and Policy 9: 110-118.
\bibAnnoteFile{Barton:2004}

\bibitem{Lorenz:2009}
Lorenz T (2009) Abductive fallacies with agent-based modelling and system
  dynamics.
\newblock Epistemological Aspects of Computer Simulation in the Social Sciences
  5466: 141-152.
\bibAnnoteFile{Lorenz:2009}

\bibitem{Borshchev:2004}
Borshchev A, Filippov A (2004) From system dynamics and discrete event to
  practical agent based modeling: Reasons, techniques, tools.
\newblock In: Proceedings of the XXII International Conference of the System
  Dynamics society.
\bibAnnoteFile{Borshchev:2004}

\bibitem{Figueredo:2012:Thesis}
Figueredo GP (2012) Translating Simulation Approaches for Immunology.
\newblock Ph.D. thesis, School of Computer Science, The University of
  Nottingham.
\bibAnnoteFile{Figueredo:2012:Thesis}

\bibitem{Forrester:1958}
Forrester JW (1958) Industrial dynamics--a major breakthrough for decision
  makers.
\newblock Harvard Business Review 36: 37-66.
\bibAnnoteFile{Forrester:1958}

\bibitem{Kirkwood:1998}
Kirkwood CW (1998) System Dynamics Methods: A Quick Introduction.
\bibAnnoteFile{Kirkwood:1998}

\bibitem{Pugh:1981}
Richardson GP, Pugh AL (1981) Introduction of System Dynamics Modelling with
  {DYNAMO}.
\newblock {MIT} Press, Cambridge, {MA}, {USA}.
\bibAnnoteFile{Pugh:1981}

\bibitem{Macal:2010}
Macal CM (2010) To agent-based simulation from system dynamics.
\newblock In: Proceedings of the 2010 Winter Simulation Conference.
\bibAnnoteFile{Macal:2010}

\bibitem{Wooldridge:2002}
Wooldridge M (2002) An Introduction to Multiagent Systems.
\newblock John {W}iley and {S}ons {I}nc, {E}ngland.
\bibAnnoteFile{Wooldridge:2002}

\bibitem{Macal:2010b}
C~M~Macal MJN (2010) Tutorial on agent-based modelling and simulation.
\newblock Journal of Simulation 4: 151-162.
\bibAnnoteFile{Macal:2010b}

\bibitem{Siebers:2007}
Siebers PO, Aickelin U (2007) Encyclopaedia of Decision Making and decision
  support technologies, chapter Introduction to Multi-Agent Simulation.
\newblock pp. 554-564.
\bibAnnoteFile{Siebers:2007}

\bibitem{Scholl:2001a}
Scholl HJ (2001) Agent-based and system dynamics modeling: a call for cross
  study and joint research.
\newblock In: Proceedings of the 34th Annual Hawaii International Conference on
  Systems Sciences.
\bibAnnoteFile{Scholl:2001a}

\bibitem{Pourdehnad:2002}
Pourdehnad J, Maani K, Sedehi H (2002) System dynamics and intelligent agent
  based simulation: where is the synergy?
\newblock In: Proceedings of the XX International Conference of the System
  Dynamics society.
\bibAnnoteFile{Pourdehnad:2002}

\bibitem{Stemate:2007}
Stemate L, Taylor I, Pasca C (2007) A comparison between system dynamics and
  agent based modeling and opportunities for cross-fertilization.
\newblock In: Proceedings of the 2007 Winter Simulation Conference. S. G.
  Henderson, B. Biller, M.-H. Hsieh, J. Shortle, J. D. Tew, and R. R. Barton.
\bibAnnoteFile{Stemate:2007}

\bibitem{Schieritz:2002}
Schieritz N (2002) Integrating system dynamics and agent-based modeling.
\newblock In: Proceedings of the XX International Conference of the System
  Dynamics society.
\bibAnnoteFile{Schieritz:2002}

\bibitem{Schieritz:2003}
Schieritz N, Milling PM (2003) Modeling the forrest or modeling the trees: A
  comparison of system dynamics and agent based simulation.
\newblock In: Proceedings of the XXI International Conference of the System
  Dynamics society.
\bibAnnoteFile{Schieritz:2003}

\bibitem{Schieritz:2003b}
Schieritz N, Gr{\"o}{$\beta$}ler A (2003) Emergent structures in supply chains:
  a study integrating agent-based and system dynamics modeling.
\newblock In: Proceedings of the XXI International Conference of the System
  Dynamics society.
\bibAnnoteFile{Schieritz:2003b}

\bibitem{Sterman:2008}
Ramandad H, Sterman J (2008) Heterogeneity and network structure in the
  dynamics of diffusion: Comparing agent-based and differential equation
  models.
\newblock Management Science 5.
\bibAnnoteFile{Sterman:2008}

\bibitem{Schieritz:2004}
Schieritz N (2004) Exploring the agent vocabulary -- emergency and evolution in
  system dynamics.
\newblock In: Proceedings of the 2004 system dynamics conference.
\bibAnnoteFile{Schieritz:2004}

\bibitem{Franceschi:2005}
Martinis MD, Franceschi C, Monti D, Ginaldi L (2005) Inflamm-ageing and
  lifelong antigenic load as major determinants of ageing rate and longevity.
\newblock FEBS 579: 2035-2039.
\bibAnnoteFile{Franceschi:2005}

\bibitem{Cossarizza1996173}
Cossarizza A, Ortolani C, Paganelli R, Barbieri D, Monti D, et~al. (1996)
  {CD}45 isoforms expression on {CD4}+ and {CD8}+ {T} cells throughout life,
  from newborns to centenarians: implications for {T} cell memory.
\newblock Mechanisms of Ageing and Development 86: 173 - 195.
\bibAnnoteFile{Cossarizza1996173}

\bibitem{Lorenzi:2008}
Lorenzi A, Patterson A, Pratt A, Jefferson M, CE Chapman~and\ F~Ponchel JI
  (2008) Determination of thymic function directly from peripheral blood: A
  validated modification to an established method.
\newblock Journal of Immunological Methods 389: 185-194.
\bibAnnoteFile{Lorenzi:2008}

\bibitem{Comans-Bitter:1997}
Comans-Bitter WM, de~Groot R, van~den Beemd R (1997) Immunophenotyping of blood
  lymphocytes in childhood. reference values for lymphocites subpopulations.
\newblock J Pediatr : 388-393.
\bibAnnoteFile{Comans-Bitter:1997}

\end{thebibliography}

\section*{Figure Legends}

{\bf Figure 1. The data set used as a look-up table for the active cells.} The data set contains the number of activated $CD4$ cells per $mm^3$ for early years taken from Comans-Bitter {\it et al.}~\cite{Comans-Bitter:1997}.

{\bf Figure 2. The system dynamics model, functions and parameters.} The table function RealActives returns the values for active T cells at a certain time. The functions h, g and s correspond to the functions defined in the mathematical model. In the diagram the round squares are the stocks, the grey straight arrows with an hourglass are the flows, the curved arrows are the information used to determine stock and flow values, the grey circles are constant values and the grey circles with a triangle inside are the variables or parameters of the system.

{\bf Figure 3. The naive T cell agent.} In the state-chart diagram the round squares represent the states, the arrows are the transitions, the black circles outlined with a smaller filled in circle inside it are the final states (death), the lozenge represents a decision of the initial state of the agent and the circles with a small triangle inside are the variables of the system.

{\bf Figure 5. Data sets (collected in \cite{Murray:2003}, \cite{Cossarizza1996173} and \cite{Lorenzi:2008}) used for validation of the naive T cell output simulation models.} In the figure, the original values were converted to percentage.

\section*{Tables}

\begin{table}[h!]
 \caption{Rate values for the mathematical model (obtained from~\cite{Murray:2003})}

 \centering
\begin{tabular}{|l|l|}
\hline
rate                 &  value(s)  \\
\hline \hline
$\lambda_t$ & $\frac{log(2)}{15.7}(year^{-1})$ \\
\hline
$\lambda_n$ & 0.22, 2.1, 0.003 \\
\hline
$\mu_n$ & 4.4 \\
\hline
$c$ & 0 (no proliferation) or $\mu_n(1+ \frac{300}{\bar{N_p}})$ \\
\hline
$\lambda_{mn}$ & 0 \\
\hline
$\mu_m$ & 0.05\\
\hline
$\lambda_{a}$ & 1 \\
\hline
\end{tabular}
 \label{Tab:RatesModelNaiveTCell}
\end{table}

\begin{table}[ht!]
 \caption{Parameters from the mathematical model and their correspondents from the SD model}

 \centering
\begin{tabular}{|l|l|}
\hline
rate                 &  correspondent  \\
\hline \hline
$s_0$ & $s0$ \\
\hline
$\lambda_t$ & $lambda\_t$ \\
\hline
$c$ & $c$ \\
\hline
$\overline{N}_p$ & $bar\_Np$\\
\hline
$\overline{s}$ & $bar\_s$\\
\hline
$b$  & n \\
\hline
$\lambda_n$ & $ProliferationRate$ \\
\hline
$\mu_n$ & $NaiveDeathRate$ \\
\hline
$\lambda_{mn}$ & $MemoryToNPRate$ \\
\hline
$\mu_{N_p}$ & $NPDeathRate$ \\
\hline
$\mu_m$ & $MemoryDeathRate$\\
\hline
$\lambda_{a}$ & $ReversionToMemoryRate$ \\
\hline
\end{tabular}
 \label{Tab:ParametersModelNaiveTCell}
\end{table}

\begin{table}[ht]
\caption{Flow calculations for the naive T cell output model}
\begin{center}
\begin{tabular}{|l|l|l|l|}
\hline
Stock                       &  Flow                         &  Expression                 & Flow formula       \\
\hline \hline
\multirow{9}{*}{\it Naive}&&&\\
                            &  $ThymusOutput$               &  $s_0e^{-\lambda_tt}s(N_p)$ & $s0()e^{-\lambda_t.time()}s()$  \\
                            &&&\\
                            \cline{2-4}
                            &&&\\
                            &  $NaiveCellsInProliferation$  &  $\lambda_nN$               & $ProliferationRate.Naive$         \\
                            &&&\\
                            \cline{2-4}
                            &&&\\
                            &  $NaiveDeath$                 &  $\mu_ng(N_p)N$             & $(NaiveDeathRate.$\\
                            &                               &                             & $g()Naive)$         \\
\hline
\hline
\multirow{6}{*}{\it NaiveProliferation} &&&\\
                            & $Proliferation$               &  $ch(N,N_p)$               & $(c \times h().$        \\
                            &
                                                       &                            & $NaiveProliferation)$ \\
                            \cline{2-4}
                            &&&\\
                            & $Death$                       &  $\mu_nN_p$                & $(NpDeathRate.$   \\
                            &                               &                            & $NaiveProliferation)$    \\
\hline
\hline
\multirow{9}{*}{\it Memory}&&&\\
                            & $MemoryToNaiveProlife-$  &  $\lambda_{mn}M$           & $(MemoryToNPRate.$        \\
                            & $ration$                              &                            & $ Memory)$         \\
                            \cline{2-4}
                            &&&\\
                            & $ReversionToMemory$           &  $\lambda_{a}A$            & $(ReversionToMemoryRate.$\\
                            &                               &                            & $ RealActives(time()))$         \\
                            \cline{2-4}
                            &&&\\
                            & $MemoryDeath$                 &  $\mu_mM$                  & $(MemoryDeathRate. $\\
                            &                               &                            & $Memory)$\\

\hline
\end{tabular}
\label{Tab:NaiveFlowCalculations}
\end{center}
\end{table}

\begin{table}[htpb]
\caption{Agents' parameters and behaviours for the naive T cell output model}
\begin{center}
\begin{tabular}{|l|l|l|l|}
\hline
State                       & Parameters                     &  Reactive behaviour   & Proactive behaviour       \\
\hline \hline

\multirow{4}{*}{\it Naive}  &  $NaiveDeathRate$              & Dies                  &                           \\
                            \cline{2-4}
                            &                                & Is produced           &                           \\
                            &                               & by thymus             &                           \\
                            \cline{2-4}
                            & $ProliferationRate$            &                       & Reproduces                \\
\hline
\hline
\multirow{7}{*}
{\it NaiveProliferation}    & $NaiveProliferationDeathRate$  & Dies                  &                           \\
                            \cline{2-4}
                            &$ProliferationRate$            & Is produced by        &                           \\
                            &                                & Naive proliferation   &                           \\
                            \cline{2-4}
                            & $MemoryToNPRate$                & Is produced from      &                           \\
                            &                                 & Memory                &                           \\
                            \cline{2-4}
                            & $NaiveProliferationRate$       &                       &  Reproduces               \\
\hline
\hline
\multirow{4}{*}
{\it Memory}                & $MemoryDeathRate$              & Dies                  &                           \\
                            \cline{2-4}
                            & $ReversionToMemoryRate$        & Is produced from      &                           \\
                            &                                & active cells          &                           \\
                            \cline{2-4}
                            & $MemoryToNPRate$               &                       & Turns into Naive          \\
\hline
\end{tabular}
\label{Tab:NaiveABSBehaviours}
\end{center}
\end{table}

\begin{table}[ht]

\caption{Simulation parameters for different scenarios. The parameter $c$ is only used in the first scenario, where there is no proliferation. In the other scenarios, proliferation is defined by the equation $\left ( 1+ \frac{300}{\bar{N_p}} \right )* \frac{4.4}{h}$~\cite{Murray:2003}.}
\begin{center}
\begin{tabular}{|l|l|c|c|c|c|c|c|c|}

\hline
Scenario                    & Description       & \multicolumn{7}{|c|}{Parameters}  \\
                            &                   &  $\lambda_n$ & $\lambda_{mn}$&$\overline{N}_p$ & $\overline{s}$ & $b$  &$\mu_{N_p}$  & $c$\\
\hline
 1 & No peripheral proliferation                &   0.22      & 0.05          & 387             & 0.48           & 3.4  & 0.13     & 0\\
\hline
 2 & No homeostatic reduction in thymic export, &   2.1       & 0             & 713             & 0              & 0    & 4.4      &-- \\
   & no homeostatic alteration of naive death rate & & & & & & & \\
\hline
 3 & Homeostatic alteration of naive death rate &   0.003     & 0             & 392             & 0              & 4.2  &  4.4     & -- \\
   & but not thymic export & & & & & & & \\
\hline
 4 & Homeostatic alteration of thymic export but&   0.005     & 0             & 378             & 2.4            & 0    & 4.4      & --\\
  & no naive death rate & & & & & & & \\
\hline
 5 & No restrictions                            &   0.005     & 0             & 378             & 2.2            & 0.13 & 4.4      & --\\
\hline
\end{tabular}
\label{Tab:ModelNaiveTCellsParameters}
\end{center}
\end{table}

\begin{table}[ht]
 \centering

 \caption{The data set used for validation obtained in \cite{Murray:2003} and \cite{Cossarizza1996173}}

\begin{tabular}{|l|l|l|}
\hline
Age                   &  $\frac{\log_{10}TREC}{{10^6 \times n} PBMC}$     & number of individuals  \\
\hline \hline
0   & 5.03 & 48 \\
\hline
1-4 & 4.93 & 53 \\
\hline
5-9 & 4.86 & 19 \\
\hline
10-14 & 4.86 & 19 \\
\hline
15-19 & 4.56 & 33 \\
\hline
20-24 & 3.88 & 26 \\
\hline
25-29 & 3.75 & 47 \\
\hline
30-34 & 3.61 & 65 \\
\hline
35-39 & 3.54 & 73 \\
\hline
40-44 & 3.52 & 52\\
\hline
45-49 & 3.37 & 55\\
\hline
50-54 & 3.17 & 16\\
\hline
\end{tabular}
 \label{Tab:datasetMurray}
\end{table}

\begin{table}[ht]
 \caption{The data set collected in Lorenzi {\it et al.} \cite{Lorenzi:2008}}

 \centering
\begin{tabular}{|l|l|l|}
\hline
Age                   &  $\frac{\log_{10}TREC}{{10^6 \times n} PBMC}$     & number of individuals  \\
\hline \hline
0   & 4.85 & 2 \\
\hline
1-4 & 5.29 & 30 \\
\hline
5-9 & 5.05 & 33 \\
\hline
10-14 & 4.99 & 15 \\
\hline
15-19 & 4.56 & 5 \\
\hline
20-24 & 4.55 & 12 \\
\hline
25-29 & 4.55 & 9 \\
\hline
30-34 & 4.44 & 20 \\
\hline
35-39 & 4.23 & 15 \\
\hline
40-44 & 4.16 & 9\\
\hline
45-49 & 3.82 & 16\\
\hline

50-54 & 4.21 &21\\
\hline
\end{tabular}
 \label{Tab:datasetLorenzi}
\end{table}

\begin{table}[!htpb]
 \caption{Wilcoxon test with $5\%$ significance level comparing the results from SDMS and ABMS}

 \centering
\begin{tabular}{|l|c|}
\hline
Scenario                  & p \\
\hline
1                         &  0.8650  \\
\hline
2                         &  0.8750  \\
\hline
3                         &  0.7987  \\
\hline
4                         &  0.8408  \\
\hline
5                         &  0.9719  \\
\hline

\end{tabular}
 \label{Tab:ModeNaiveTCellsWilcoxon}
\end{table}

\end{document}